\documentclass{article}
\pdfoutput=1

\usepackage[final, nonatbib]{neurips_2020}
\usepackage[utf8]{inputenc} % allow utf-8 input
\usepackage[T1]{fontenc}    % use 8-bit T1 fonts
\usepackage{hyperref}       % hyperlinks
\usepackage{url}            % simple URL typesetting
\usepackage{booktabs}       % professional-quality tables
\usepackage{amsfonts}       % blackboard math symbols
\usepackage{nicefrac}       % compact symbols for 1/2, etc.
\usepackage{microtype}      % microtypography
\usepackage{verbatim}

\usepackage[sort, numbers]{natbib}
\usepackage{xcolor}
\definecolor{navyblue}{rgb}{0.0, 0.0, 0.5}
\hypersetup{colorlinks, citecolor=blue}
\usepackage{graphicx}

\usepackage{enumitem}

\usepackage{algorithmic}
\usepackage{algorithm}

\usepackage{amssymb}
\usepackage{amsmath}
\usepackage{amsthm}
\usepackage{bbm}
\usepackage{wrapfig,lipsum,booktabs}

\usepackage{multirow}
\usepackage{multicol}
\usepackage{subcaption}
\usepackage{caption, makecell}
\usepackage{array}

\usepackage{CJKutf8}

\usepackage{comment}

\makeatletter
\def\thanks#1{\protected@xdef\@thanks{\@thanks
        \protect\footnotetext{#1}}}
\makeatother

\usepackage{ifpdf}
\title{MixCo: Mix-up Contrastive Learning for \\Visual Representation}

\author{Sungnyun Kim*,
  Gihun Lee*,
  Sangmin Bae* \vspace{-10pt} \AND Se-Young Yun \\
  KAIST\\
    \texttt{\{ksn4397, opcrisis, bsmn0223, yunseyoung\}@kaist.ac.kr}
    \thanks{* Equal contribution. The names are randomly ordered.}
}
\vspace{-10pt}
\begin{document}
\begin{CJK}{UTF8}{mj} %% 한글 나오게 하기 위한 부분
\maketitle
\vspace{-10pt}
\begin{abstract}
Contrastive learning has shown remarkable results in recent self-supervised approaches for visual representation. By learning to contrast positive pairs' representation from the corresponding negatives pairs, one can train good visual representations without human annotations. This paper proposes Mix-up Contrast (\textbf{MixCo}), which extends the contrastive learning concept to \textit{semi-positives} encoded from the mix-up of positive and negative images. MixCo aims to learn the relative similarity of representations, reflecting how much the mixed images have the original positives. We validate the efficacy of MixCo when applied to the recent self-supervised learning algorithms under the standard linear evaluation protocol on TinyImageNet, CIFAR10, and CIFAR100. In the experiments, MixCo consistently improves test accuracy. Remarkably, the improvement is more significant when the learning capacity (e.g., model size) is limited, suggesting that MixCo might be more useful in real-world scenarios. The code is available at: \url{https://github.com/Lee-Gihun/MixCo-Mixup-Contrast}.
\vspace{-5pt}
\end{abstract}
\vspace{-5pt}
\section{Introduction}
\vspace{-4pt}
\begin{wrapfigure}{R}{0.45\textwidth}
\centering
\includegraphics[width=0.45\textwidth]{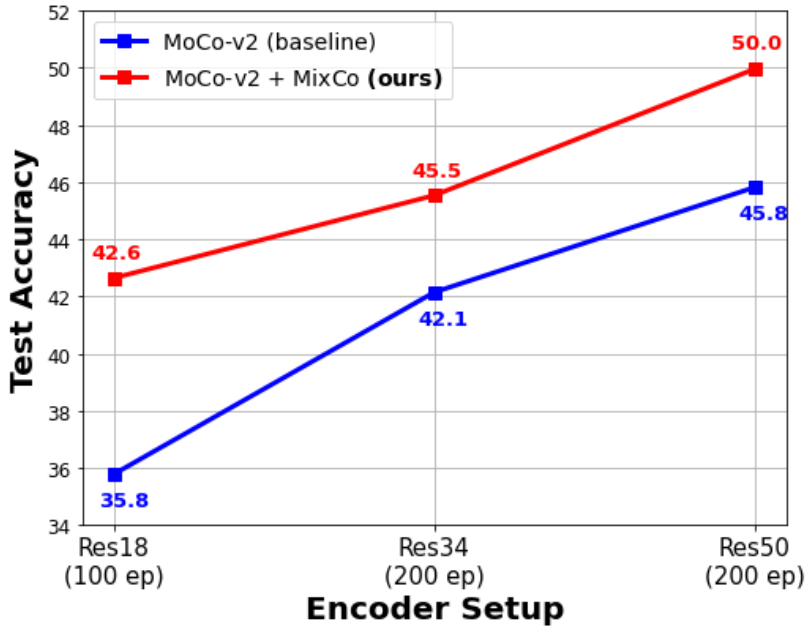}
\caption{\label{fig:mixco_arch} Performance of MixCo on TinyImageNet (linear-evaluation) of different model sizes.}
\vspace{-10pt}
\end{wrapfigure}

Learning representations without human annotations on data is very promising since it enables us to utilize a massive amount of data without demanding effort to label them \citep{unsupervised}. Self-supervised learning is an approach that exploits supervised signals through pretext (\textit{or proxy}) tasks \citep{ssl_survey}. After the remarkable successes in natural language processing \citep{bert, gpt2, gpt3}, recent studies are closing the gap with supervised learning in computer vision \citep{moco2, byol, simclr2, swav}. Nowadays, the critical component of state-of-the-art achievements in self-supervised visual representation learning is contrastive learning \citep{cpc}, which learns to discriminate positive pairs from negative pairs. More specifically, it learns to embed the representations of differently augmented versions of the same image (\textit{positive pair}) to be similar, while to be dissimilar if they came from different images (\textit{negative pair}). Such representations are highly transferable to down-stream tasks \citep{moco, simclr, hard_negative}, and even outperform the representations learned by the supervised manner in some tasks \citep{moco2, what_makes_discrimination_good}. However, the performance of learning is highly sensitive to the choice of augmentations \citep{simclr} and requires a number of negatives \citep{simclr, moco}.

In this paper, inspired by the success of mix-up training \citep{mixup} in recent studies \citep{mixup_manifold, on_mixup_training, rethinking_mixup_unsupervised}, we propose Mix-up Contrast \textbf{(MixCo)} as an extension of contrastive learning approach. MixCo generates the representation from a mixed image and learns how much the representation should be similar to those from the different views of each image used in the mix-up. Here, mixing images is a process of generating semi-positive pairs since the images before mixing are originally negative pairs of each other. We argue that learning through such semi-positive samples helps to learn better representations by preventing the discrimination of positive from being trivial. 

Our method is simple yet effective and can be applied on-the-fly to various contrast learning based methods. To validate the efficacy of MixCo, it is applied to recent popular representation learning methods: MoCo-v2 \citep{moco2} and SimCLR \citep{simclr}. We show that MixCo consistently improves the linear classification results when transferred to three different datasets: CIFAR10 \citep{cifar}, CIFAR100 \citep{cifar}, and TinyImageNet \citep{tiny-imagenet}. Notably, the improvement is more significant when the training resource is more limited. Our contributions are: (\textit{i}) We extended the concept of contrastive loss for visual representation learning to semi-positive samples. (\textit{ii}) We introduced MixCo, an on-the-fly method to exploit semi-positive samples based on the popular mix-up training approach. (\textit{iii}) We show that MixCo enables to use of the given training samples more effectively, especially when the computational budget is limited.
\vspace{-5pt}
\section{Related Work}
\vspace{-4pt}
\paragraph{Contrastive Learning} Although earlier self-supervised methods handcrafted pretext tasks \citep{pretext_colorization,pretext_context,pretext_jigsaw,pretext_moving,pretext_spot,pretext_transitive}, most recent approaches that show remarkable performance are based on instance-level classification with contrastive loss \citep{deepinfomax, simclr, simclr2, moco, moco2, cpc, cmc}. The contrastive loss \citep{cpc} measures the similarity of representation pairs and tries to \textit{contrast} the positive and negative pairs. Since comparing all the possible pairs are computationally prohibitive, most methods relax the problem as discrimination to randomly sampled subsets \citep{swav}. In practice, the performance relies on careful settings on how to form negative pairs and the choice of augmentation strategies \citep{simclr, simclr2}. To overcome the dependency of the negative sample size on batch size, MoCo \citep{moco, moco2} preserved the representations of samples as a queue. SwAV \citep{swav} combined the simultaneous clustering algorithm and conducted swapped prediction to enforce consistent mapping between views.

\paragraph{Mix-up Training} \citep{mixup} proposed mix-up training as a method to regularize the training of the network by using the samples from the vicinity of data. The vicinity samples and their labels are generated by a convex combination of samples from different classes. Mix-up training consistently improved the performance in various studies \citep{mixup_manifold, on_mixup_training}. Recently, \citep{rethinking_mixup_unsupervised} applied mix-up training to obtain augmented views of images, but only the input data are mixed-up instead of considering the desirable similarity. In \citep{hard_negative}, the mix-up is used to generate hard negative samples so that discriminating positive pairs become more challenging.
\vspace{-5pt}
\section{Method}
\vspace{-4pt}

\paragraph{Contrastive Loss} Contrastive (or InfoNCE \citep{cpc}) loss tries to maximize the similarity of positive pairs and minimizes that of negative pairs:

\begin{equation}
\mathcal{L}_{Contrast} = -\sum_{i=1}^n \log\frac{\exp(\frac{v_i \cdot v_{i}'}{ \tau})}{\sum_{j=0}^{r} \exp(\frac{v_i \cdot v_{j}'}{\tau})}
\label{eq:Contrast}
\end{equation}

where $n$ is the batch size, $\tau$ is a temperature to soften the similarity logits, and $v$ is the normalized representation vector. Note that we denote the representation from view1 as $v$ and from view2 as $v'$. Each is referred to as query and key in MoCo \citep{moco, moco2}. Note that the loss function is the same with ($r$+1)-way softmax function for one positive and $r$ number of negative classes. Although the methods based on the contrastive learning showed impressive results \citep{cpc, moco2, simclr2}, learning to discriminate the positive from negatives is likely to be overfitted. Since it relies heavily on discriminating instance-level semantics, once the model can discriminate the positive from the negatives, there is not enough chance to learn from the negatives. The contrastive loss equally pushes the given representation away from that of negatives. Applying adequate augmentations helps, but as \citep{simclr} suggested, such augmentations are very sensitive to the combinations.

\begin{figure}[!t]
    \centering
    \begin{subfigure}[b]{0.48\textwidth}
        \includegraphics[width=\textwidth]{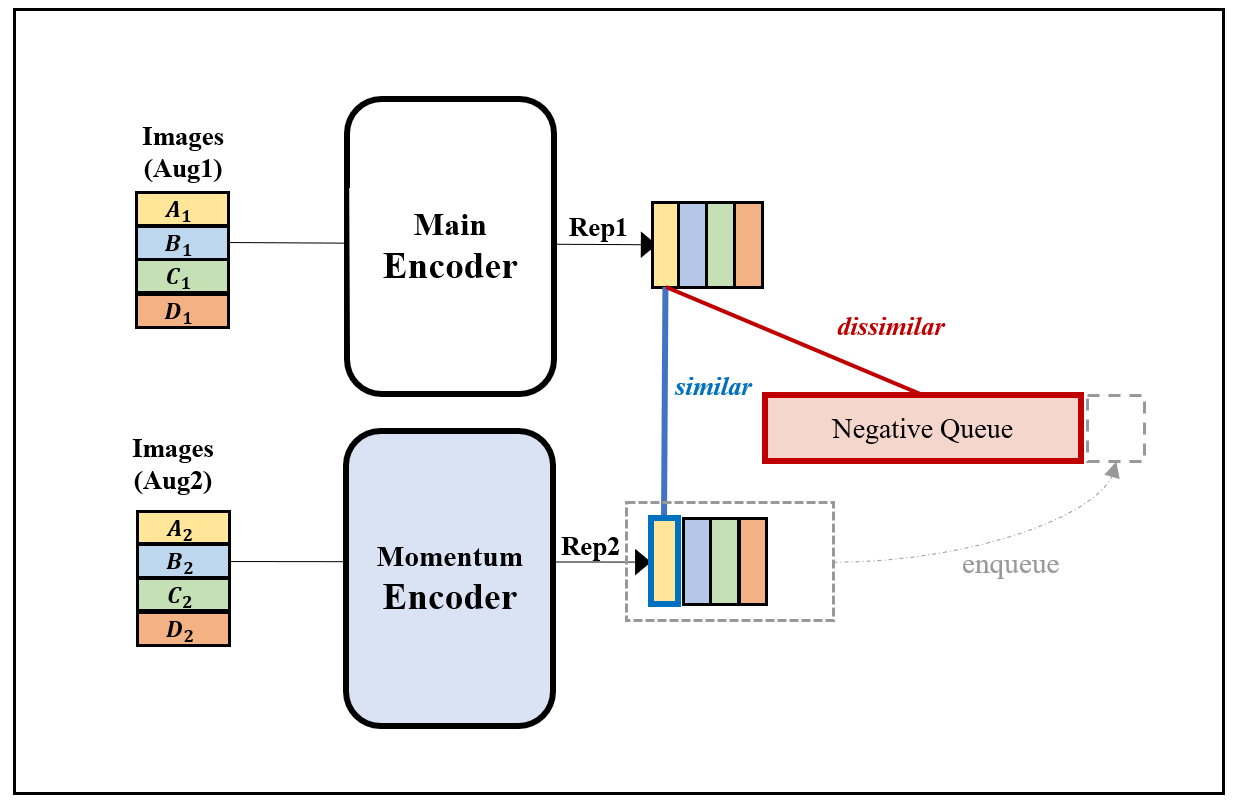}
        \caption{MoCo (Momentum Contrast)}
        \label{subfig:moco_overview}
    \end{subfigure}
    \hspace{10pt}
    \begin{subfigure}[b]{0.48\textwidth}
        \includegraphics[width=\textwidth]{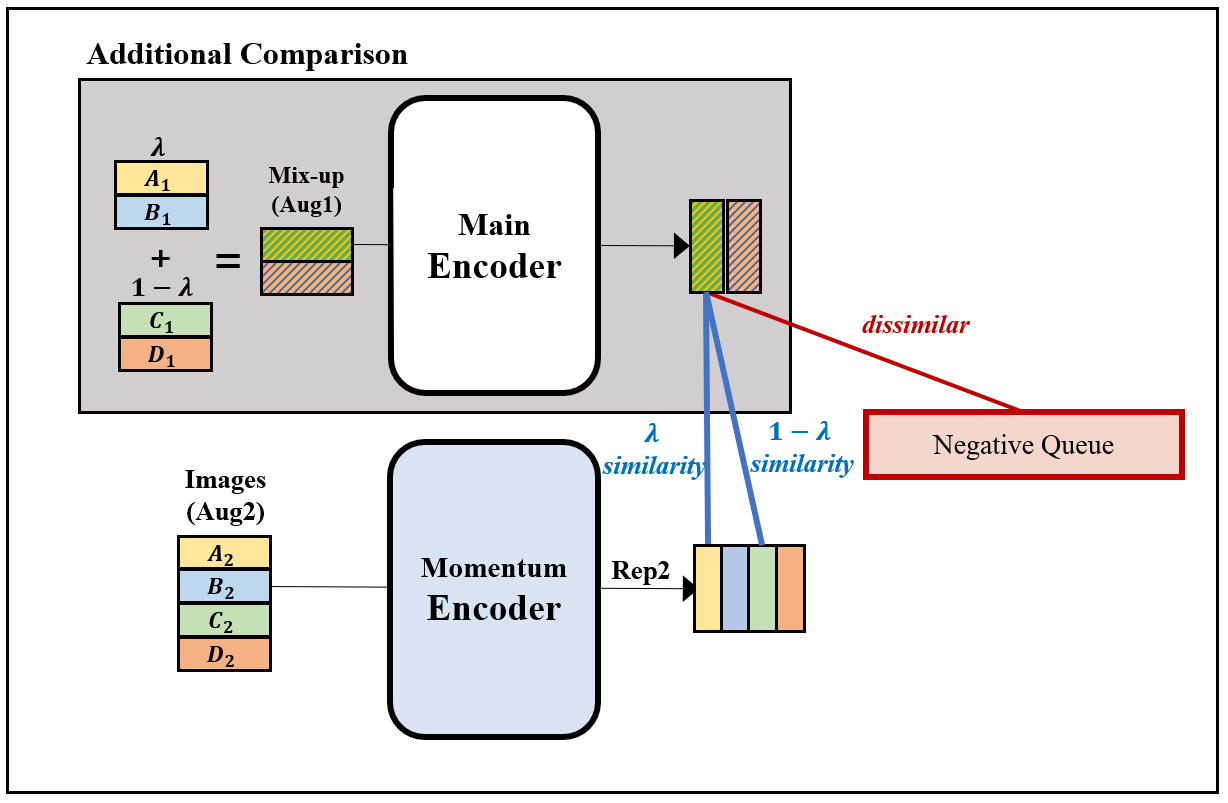}
        \caption{MixCo (Mix-up Contrast)}
        \label{subfig:mixco_overview}
    \end{subfigure}
    \caption[MixCo overview]{An overview of MoCo and MixCo. (a) The representations are compared with their counterparts from the different augmentation strategy. (b) The representation from mixed-up images are \emph{semi-positive} with the representations of the original images (different augmentation). The negatives are shared with the original MoCo queue. See Appendix \ref{appendix:algorithm} for the detailed algorithm.}
    \label{fig:overview}
\end{figure}

\paragraph{Mix-up Contrast}

The motivation of MixCo is exploiting softened data and targets for the contrastive learning to relieve the instance discrimination problem by letting the model learn the implicit relationship between the positives and the negatives. Similar to mix-up training, MixCo generates mixed data $x_{{mix}_{i,k}}$ using convex combination between data $x_i$ and data $x_k$, and its representation $v_i^{{mix}_{i,k}}$ is defined as follows:
\vspace{5pt}
\begin{equation}
       x_{mix_{i,k}} = \lambda_i \cdot x_{i} + (1-\lambda_i) \cdot x_{k}, \qquad \quad v_i^{{mix}_{i,k}} = {f}_{\text{encoder}}(x_{mix_{i,k}}) 
\end{equation}
\vspace{5pt}
where $k$ represents any arbitrary data index in the same batch with $x_i$. Now we have 2 semi-positive pairs of which each corresponds to data $x_{i}$ and data $x_{k}$. By letting the desired similarity to ${v}_{i}'$ and ${v}_{k}'$ as $\lambda_i$ and $1-\lambda_i$, respectively, we have:
%\begin{equation*}
%    {L}_{MixCo} = -\sum_{i=1}^a \Bigg[log\Big(\frac{\lambda_i \cdot exp(v_{i}^{mix_{i,i+a}} \cdot v_{i}^{'} / \tau_{mix}) + (1 - \lambda_{i}) \cdot exp(v_{i}^{mix_{i,i+a}} \cdot v_{i+a}^{'} / \tau_{mix})}{\sum_{k=0}^{K} exp(v_{i}^{mix_{i,i+a}} \cdot v_{k}^{'} / \tau_{mix})}\Big) \Bigg]
%\label{eq:mixco}
%\end{equation*}
\vspace{5pt}
\begin{equation}
    \small
    \mathcal{L}_{MixCo} = 
    -\sum_{i=1}^{n_{mix}} 
    \Bigg[
    \lambda_i\cdot 
    \log\underbrace{\Bigg(\frac{\exp\Big(\frac{v_i^{mix_{i,k}} 
    \cdot v'_i}{\tau_{mix}}\Big)}
    {\sum_{j=0}^{r} \exp\Big(\frac{v_i^{mix_{i,k}} \cdot v'_j}{\tau_{mix}}\Big)} \Bigg)}_{\text{Similarity to $x_i$}}  
    + 
    (1-\lambda_i) \cdot 
    \log\underbrace{\Bigg(\frac{\exp\Big(\frac{v_i^{mix_{i,k}} \cdot v'_k} {\tau_{mix}}\Big)} 
    {\sum_{j=0}^{r} \exp\Big(\frac{v_i^{mix_{i,k}} \cdot v'_j}{\tau_{mix}}\Big)} \Bigg)}_{\text{Similarity to $x_k$}} 
    \Bigg]
\label{eq:mixco}
\end{equation}
\vspace{5pt}
where $n_{mixco}$ is the number of mixed samples to be used at MixCo and $\tau_{mix}$ is softening temperature for similarity logits. Finally, the total loss is defined as a combination of $\mathcal{L}_{Contrast}$ and $\mathcal{L}_{MixCo}$ using a hyperparameter $\beta$:
\vspace{5pt}
\begin{equation}
 \mathcal{L}_{total} = \mathcal{L}_{Contrast} + \beta\mathcal{L}_{MixCo}
\end{equation}
\vspace{5pt}
Note that $x_i$ and $x_k$ are negatives (i.e., does not share the semantics) for each other. Therefore, the mixed image representation $v_i^{{mix}_{i,k}}$ has \textit{semi-positive} relation to the representations $v_i$ and $v_k$, which are from the original images. Learning to capture such a relation is much harder than merely discriminating against a single positive from the negatives, which leads to more efficient use of negatives. An example of MixCo application to MoCo \citep{moco, moco2} is illustrated in  \autoref{fig:overview}. 
\vspace{-5pt}
\section{Experiment}
\vspace{-4pt}

\subsection{Experimental setup}
\vspace{-3pt}
In unsupervised learning, we train ResNet \citep{resnet} encoder with different depths on TinyImageNet. ResNet-34 and ResNet-50 encoder are trained for 200 epochs with batch size 128, and ResNet-18 is trained with 100 epochs. For MixCo hyperparameters, $\beta=1$ and $\tau_{mix}=0.05$. For linear evaluation, we follow the standard protocol, training a classifier for 100 epochs with initial learning rate 3.0 while freezing the pretrained encoder. Both in pretraining and linear evaluation, other unspecified hyperparameters are the same as MoCo-v2 \citep{moco2}, including augmentations, optimizer, and learning rate scheduler. SimCLR results are yet to be provided with thorough hyperparameter search.

\subsection{Linear evaluation results}
\vspace{-3pt}
\autoref{tab:linear_eval} shows the overall linear evaluation results on different pretrain setups. All the encoders are pretrained on TinyImageNet \citep{tiny-imagenet}, and the linear classifier is trained to be transferred on different datasets. In the experiment, MixCo consistently improves on different encoder sizes. Note that the encoders do not see the CIFAR10 \citep{cifar} and CIFAR100 \citep{cifar} datasets in the pretraining phase. MixCo also improves the results on such datasets, suggesting that the obtained embedding encoded by MixCo is not limited to a specific dataset used at the pretraining phase. The improvement by applying MixCo is most significant ($+6.84\%$) at ResNet-18 with 100 epochs, where the smallest training resource is used. The comparison of the required training resources for pretraining is in Appendix \ref{appendix:computation}. Additional experiments, including ImageNet dataset and pretraining with longer epochs, are found in Appendix \ref{appendix:additional}.
\begin{table}[h!]
\vspace{-10pt}
\footnotesize
  \caption{Linear evaluation results on TinyImageNet \citep{tiny-imagenet} pretrained encoders.}
  \vspace{3pt}
  \label{tab:linear_eval}
  \centering
  \begin{tabular}{ccc|ccc}
    \toprule
    \multirow{2}{*}{Architecture (epochs)}      &   \multirow{2}{*}{Method}     & \multirow{2}{*}{MixCo (ours)}     &   \multicolumn{3}{c}{Top-1 Accuracy (\%)} \\                                                                                                 &                               &                                   &   TinyImageNet    &   CIFAR10     & CIFAR100  \\
    \midrule\midrule

    \multirow{3}{*}{ResNet-18 (100)}            &   Supervised                  &   -                               &   63.96               &   -                   &   -               \\
                                                &   \multirow{2}{*}{MoCo-v2 \citep{moco2}}       &   X               &   35.79               &   71.02               &   48.81           \\ 
                                                &                                               &   O               &   \textbf{42.65}      &   \textbf{74.25}      &   \textbf{53.23}  \\
                                                &   \multirow{2}{*}{SimCLR \citep{simclr}}       &   X               &   35.30               &   70.64               &   47.48           \\ 
                                                &                                               &   O               &   \textbf{36.44}      &   \textbf{71.44}      &   \textbf{48.32}  \\
                                                %&   \multirow{2}{*}{SwAV\citep{swav}}           &   X               &   48.80              &   0                   &   0               \\ 
                                                %&                                               &   O               &   \textbf{0}         &   \textbf{0}          &   \textbf{0}      \\
    \midrule
    \multirow{3}{*}{ResNet-34 (200)}            &   Supervised                                  &   -               &   64.99               &   -                   &   -               \\
                                                &   \multirow{2}{*}{MoCo-v2 \citep{moco2}}       &   X               &   42.15               &   72.97               &   48.56           \\ 
                                                &                                               &   O               &   \textbf{45.54}      &   \textbf{75.06}      &   \textbf{52.67}  \\
    \midrule
    \multirow{3}{*}{ResNet-50 (200)}            &   Supervised                                  &   -               &   67.03               &   -                   &   -               \\
                                                &   \multirow{2}{*}{MoCo-v2 \citep{moco2}}       &   X               &   45.82               &   78.57               &   58.21           \\ 
                                                &                                               &   O               &   \textbf{49.96}      &   \textbf{80.60}      &   \textbf{59.36}  \\
    \bottomrule
  \end{tabular}
  \vspace{-10pt}
\end{table}
\subsection{Visualization of learned representations}
\begin{figure}[htbp]
\vspace{-10pt}
    \centering
    \begin{subfigure}[b]{0.45\textwidth}
        \includegraphics[width=\textwidth]{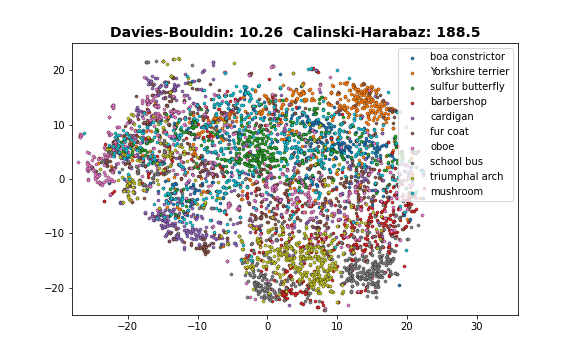}
        \caption{MoCo-v2 ResNet-18 Encoder (baseline)}
        \label{subfig:moco_tsne_tiny}
    \end{subfigure}
    \hspace{-20pt}
    \begin{subfigure}[b]{0.45\textwidth}
        \includegraphics[width=\textwidth]{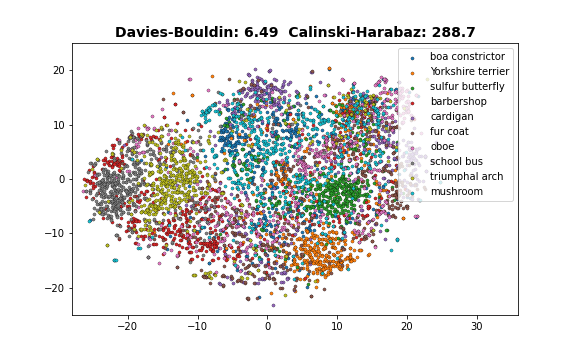}
        \caption{MixCo ResNet-18 Encoder (ours)}
        \label{subfig:mixco_tsne_tiny}
    \end{subfigure}
    \caption[MixCo]{Visualization of learned train set representations. Each encoders are pretrained on TinyImageNet. 10000 training samples from 10 different classes are used. The classes are randomly chosen to have different categories (ex. terrier, coat, bus). The evaluation metrics are at the title of each figure.}
    \label{fig:tsne_tiny}
    \vspace{-5pt}
\end{figure}

\vspace{-3pt}
The t-SNE\citep {tsne} visualization of our learned representation is in \autoref{fig:tsne_tiny}.  As shown in \autoref{subfig:mixco_tsne_tiny}, MixCo captures the semantics of data better than the baseline. We evaluate each result using Davies-Bouldin index \citep{davies_cluster} and Calinski-Harabasz index \citep{calinski_cluster}. Note that in the Davies-Bouldin index, the lower value indicates the better result, whereas the higher value is better in the Calinski-Harabasz index.

\vspace{-10pt}
\section{Conclusion}
\vspace{-4pt}

We introduce MixCo, a novel method applicable to recent self-supervised learning algorithms as on-the-fly, where contrastive learning is the key component. MixCo extends the concept of contrastive learning, which discriminates positive pairs from their negatives, to semi-positive pairs generated by the mix-up. By learning to capture the relative similarities for semi-positive pairs, the representations learned by MixCo show improved performance when transferred to downstream tasks. The improvement is more significant when training resource is limited, which implies the potential usefulness in real-world situations.

\clearpage
\section*{Acknowledgments}
This research was supported by SK Hynix AICC (K0\_efficient\_unsupervised\_representation\_learning).

\bibliography{references}

\begin{thebibliography}{10}

\bibitem{pretext_moving}
Pulkit Agrawal, Joao Carreira, and Jitendra Malik.
\newblock Learning to see by moving.
\newblock In {\em Proceedings of the IEEE international conference on computer
  vision}, pages 37--45, 2015.

\bibitem{gpt3}
Tom~B Brown, Benjamin Mann, Nick Ryder, Melanie Subbiah, Jared Kaplan, Prafulla
  Dhariwal, Arvind Neelakantan, Pranav Shyam, Girish Sastry, Amanda Askell,
  et~al.
\newblock Language models are few-shot learners.
\newblock {\em arXiv preprint arXiv:2005.14165}, 2020.

\bibitem{calinski_cluster}
Tadeusz Cali{\'n}ski and Jerzy Harabasz.
\newblock A dendrite method for cluster analysis.
\newblock {\em Communications in Statistics-theory and Methods}, 3(1):1--27,
  1974.

\bibitem{swav}
Mathilde Caron, Ishan Misra, Julien Mairal, Priya Goyal, Piotr Bojanowski, and
  Armand Joulin.
\newblock Unsupervised learning of visual features by contrasting cluster
  assignments.
\newblock {\em arXiv preprint arXiv:2006.09882}, 2020.

\bibitem{simclr}
Ting Chen, Simon Kornblith, Mohammad Norouzi, and Geoffrey Hinton.
\newblock A simple framework for contrastive learning of visual
  representations.
\newblock {\em arXiv preprint arXiv:2002.05709}, 2020.

\bibitem{simclr2}
Ting Chen, Simon Kornblith, Kevin Swersky, Mohammad Norouzi, and Geoffrey
  Hinton.
\newblock Big self-supervised models are strong semi-supervised learners.
\newblock {\em arXiv preprint arXiv:2006.10029}, 2020.

\bibitem{moco2}
Xinlei Chen, Haoqi Fan, Ross Girshick, and Kaiming He.
\newblock Improved baselines with momentum contrastive learning.
\newblock {\em arXiv preprint arXiv:2003.04297}, 2020.

\bibitem{davies_cluster}
David~L Davies and Donald~W Bouldin.
\newblock A cluster separation measure.
\newblock {\em IEEE transactions on pattern analysis and machine intelligence},
  (2):224--227, 1979.

\bibitem{bert}
Jacob Devlin, Ming-Wei Chang, Kenton Lee, and Kristina Toutanova.
\newblock Bert: Pre-training of deep bidirectional transformers for language
  understanding.
\newblock {\em arXiv preprint arXiv:1810.04805}, 2018.

\bibitem{pretext_context}
Carl Doersch, Abhinav Gupta, and Alexei~A Efros.
\newblock Unsupervised visual representation learning by context prediction.
\newblock In {\em Proceedings of the IEEE international conference on computer
  vision}, pages 1422--1430, 2015.

\bibitem{byol}
Jean-Bastien Grill, Florian Strub, Florent Altch{\'e}, Corentin Tallec,
  Pierre~H Richemond, Elena Buchatskaya, Carl Doersch, Bernardo~Avila Pires,
  Zhaohan~Daniel Guo, Mohammad~Gheshlaghi Azar, et~al.
\newblock Bootstrap your own latent: A new approach to self-supervised
  learning.
\newblock {\em arXiv preprint arXiv:2006.07733}, 2020.

\bibitem{moco}
Kaiming He, Haoqi Fan, Yuxin Wu, Saining Xie, and Ross Girshick.
\newblock Momentum contrast for unsupervised visual representation learning.
\newblock {\em arXiv preprint arXiv:1911.05722}, 2019.

\bibitem{resnet}
Kaiming He, Xiangyu Zhang, Shaoqing Ren, and Jian Sun.
\newblock Deep residual learning for image recognition.
\newblock In {\em Proceedings of the IEEE conference on computer vision and
  pattern recognition}, pages 770--778, 2016.

\bibitem{cpc2}
Olivier~J H{\'e}naff, Aravind Srinivas, Jeffrey De~Fauw, Ali Razavi, Carl
  Doersch, SM~Eslami, and Aaron van~den Oord.
\newblock Data-efficient image recognition with contrastive predictive coding.
\newblock {\em arXiv preprint arXiv:1905.09272}, 2019.

\bibitem{unsupervised}
Geoffrey~E Hinton, Terrence~Joseph Sejnowski, Tomaso~A Poggio, et~al.
\newblock {\em Unsupervised learning: foundations of neural computation}.
\newblock MIT press, 1999.

\bibitem{deepinfomax}
R~Devon Hjelm, Alex Fedorov, Samuel Lavoie-Marchildon, Karan Grewal, Phil
  Bachman, Adam Trischler, and Yoshua Bengio.
\newblock Learning deep representations by mutual information estimation and
  maximization.
\newblock {\em arXiv preprint arXiv:1808.06670}, 2018.

\bibitem{pretext_spot}
Simon Jenni and Paolo Favaro.
\newblock Self-supervised feature learning by learning to spot artifacts.
\newblock In {\em Proceedings of the IEEE Conference on Computer Vision and
  Pattern Recognition}, pages 2733--2742, 2018.

\bibitem{ssl_survey}
Longlong Jing and Yingli Tian.
\newblock Self-supervised visual feature learning with deep neural networks: A
  survey.
\newblock {\em IEEE Transactions on Pattern Analysis and Machine Intelligence},
  2020.

\bibitem{hard_negative}
Yannis Kalantidis, Mert~Bulent Sariyildiz, Noe Pion, Philippe Weinzaepfel, and
  Diane Larlus.
\newblock Hard negative mixing for contrastive learning, 2020.

\bibitem{pretext_jigsaw}
Dahun Kim, Donghyeon Cho, Donggeun Yoo, and In~So Kweon.
\newblock Learning image representations by completing damaged jigsaw puzzles.
\newblock In {\em 2018 IEEE Winter Conference on Applications of Computer
  Vision (WACV)}, pages 793--802. IEEE, 2018.

\bibitem{cifar}
Alex Krizhevsky, Geoffrey Hinton, et~al.
\newblock Learning multiple layers of features from tiny images.
\newblock 2009.

\bibitem{pretext_colorization}
Gustav Larsson, Michael Maire, and Gregory Shakhnarovich.
\newblock Learning representations for automatic colorization.
\newblock In {\em European conference on computer vision}, pages 577--593.
  Springer, 2016.

\bibitem{tiny-imagenet}
Ya~Le and Xuan Yang.
\newblock Tiny imagenet visual recognition challenge.
\newblock {\em CS 231N}, 7, 2015.

\bibitem{pcl}
Junnan Li, Pan Zhou, Caiming Xiong, Richard Socher, and Steven~CH Hoi.
\newblock Prototypical contrastive learning of unsupervised representations.
\newblock {\em arXiv preprint arXiv:2005.04966}, 2020.

\bibitem{tsne}
Laurens van~der Maaten and Geoffrey Hinton.
\newblock Visualizing data using t-sne.
\newblock {\em Journal of machine learning research}, 9(Nov):2579--2605, 2008.

\bibitem{cpc}
Aaron van~den Oord, Yazhe Li, and Oriol Vinyals.
\newblock Representation learning with contrastive predictive coding.
\newblock {\em arXiv preprint arXiv:1807.03748}, 2018.

\bibitem{gpt2}
Alec Radford, Jeffrey Wu, Rewon Child, David Luan, Dario Amodei, and Ilya
  Sutskever.
\newblock Language models are unsupervised multitask learners.
\newblock {\em OpenAI Blog}, 1(8):9, 2019.

\bibitem{ILSVRC15}
Olga Russakovsky, Jia Deng, Hao Su, Jonathan Krause, Sanjeev Satheesh, Sean Ma,
  Zhiheng Huang, Andrej Karpathy, Aditya Khosla, Michael Bernstein,
  Alexander~C. Berg, and Li~Fei-Fei.
\newblock {ImageNet Large Scale Visual Recognition Challenge}.
\newblock {\em International Journal of Computer Vision (IJCV)},
  115(3):211--252, 2015.

\bibitem{rethinking_mixup_unsupervised}
Zhiqiang Shen, Zechun Liu, Zhuang Liu, Marios Savvides, and Trevor Darrell.
\newblock Rethinking image mixture for unsupervised visual representation
  learning.
\newblock {\em arXiv preprint arXiv:2003.05438}, 2020.

\bibitem{on_mixup_training}
Sunil Thulasidasan, Gopinath Chennupati, Jeff~A Bilmes, Tanmoy Bhattacharya,
  and Sarah Michalak.
\newblock On mixup training: Improved calibration and predictive uncertainty
  for deep neural networks.
\newblock In {\em Advances in Neural Information Processing Systems}, pages
  13888--13899, 2019.

\bibitem{cmc}
Yonglong Tian, Dilip Krishnan, and Phillip Isola.
\newblock Contrastive multiview coding.
\newblock {\em arXiv preprint arXiv:1906.05849}, 2019.

\bibitem{mixup_manifold}
Vikas Verma, Alex Lamb, Christopher Beckham, Amir Najafi, Ioannis Mitliagkas,
  David Lopez-Paz, and Yoshua Bengio.
\newblock Manifold mixup: Better representations by interpolating hidden
  states.
\newblock In {\em International Conference on Machine Learning}, pages
  6438--6447. PMLR, 2019.

\bibitem{pretext_transitive}
Xiaolong Wang, Kaiming He, and Abhinav Gupta.
\newblock Transitive invariance for self-supervised visual representation
  learning.
\newblock In {\em Proceedings of the IEEE international conference on computer
  vision}, pages 1329--1338, 2017.

\bibitem{mixup}
Hongyi Zhang, Moustapha Cisse, Yann~N Dauphin, and David Lopez-Paz.
\newblock mixup: Beyond empirical risk minimization.
\newblock {\em arXiv preprint arXiv:1710.09412}, 2017.

\bibitem{what_makes_discrimination_good}
Nanxuan Zhao, Zhirong Wu, Rynson~WH Lau, and Stephen Lin.
\newblock What makes instance discrimination good for transfer learning?
\newblock {\em arXiv preprint arXiv:2006.06606}, 2020.

\bibitem{eqco}
Benjin Zhu, Jun qiang Huang, Zeming Li, Xiangyu Zhang, and J.~Sun.
\newblock Eqco: Equivalent rules for self-supervised contrastive learning.
\newblock {\em ArXiv}, abs/2010.01929, 2020.

\end{thebibliography}
\bibliographystyle{plain}

\appendix
\clearpage

\def\HiLi{\leavevmode\rlap{\hbox to \hsize{\color{yellow!50}\leaders\hrule height .8\baselineskip depth .5ex\hfill}}}

\section{Algorithm}
\label{appendix:algorithm}

\vspace{-10pt}
\begin{algorithm}[H]
    \caption{Psuedocode of MixCo on MoCo \citep{moco} in PyTorch-like style.}
    \textbf{Input}: encoder $f_q$, $f_k$, training dataset $X$, batch size $B$, feature dimension $C$, queue length $K$, momentum $m$, temperature $\tau$ and $\tau_{mix}$
    \medskip
      
    $f_k$.params = $f_q$.params \quad \textcolor{teal}{// initialize key encoder networks} \par
    queue = torch.randn($C$, $K$) \quad \textcolor{teal}{// initialize a queue of K keys: [$C$x$K$]} \par
    \medskip
    
    \textbf{for} $x$ in Dataloader($X$): \quad \textcolor{teal}{// load a minibatch x with $B$ samples} \par
    \qquad $x_q$, $x_k$ = $aug$($x$), $aug$($x$) \quad  \textcolor{teal}{// randomly augmented version for query and key} \par
    \medskip
    
    \HiLi{\qquad $\lambda$ $\sim$ $Unif$(0, 1)} \par %\quad \textcolor{teal}{// sampling from uniform(0, 1) distribution}
    \HiLi{\qquad $x_q^{mix}$ = $\lambda$ * $x_{q}$[:$B$/2] + (1 - $\lambda$) * $x_{q}$[$B$/2:]} \par %\quad \textcolor{teal}{// mixup on the augmented versions for query} \par
    \medskip
    
    \HiLi{\qquad $q$, $q_{mix}$ = $f_q$.forward($x_q$), $f_q$.forward($x_q^{mix})$ \quad \textcolor{teal}{// queries: $B$x$C$, mixup queries: [($B$/2)x$C$]}} \par
    \qquad $k$ = $f_k$.forward($x_k$).detach() \quad \textcolor{teal}{// keys: $B$x$C$, no back-propagation on $f_k$} \par
    \medskip
    
    \qquad \textcolor{teal}{// MoCo update} \par
    \qquad $l_{pos}$ = torch.bmm($q$.view($B$,1,$C$), $k$.view($B$,$C$,1)) \quad \textcolor{teal}{// positive logits: [$B$x1]} \par
    \qquad $l_{neg}$ = torch.mm($q$.view($B$,$C$), queue.view($C$,$K$)) \quad \textcolor{teal}{// negative logits: [$B$x$K$]} \par
    \medskip
    
    \qquad logits = torch.cat([$l_{pos}$, $l_{neg}$], dim=1) \quad \textcolor{teal}{// logits: [$B$x(1+$K$)]} \par
    \qquad labels = torch.zeros($B$) \quad \textcolor{teal}{// all positive logits are the 0-th, labels: [$B$]} \par
    \qquad loss = torch.nn.CrossEntropyLoss(logits / $\tau$, labels) \par
    \medskip
        
    \HiLi{\qquad \textcolor{teal}{// MixCo update}} \par
    \HiLi{\qquad $l_{semi\_pos}$ = torch.mm($q_{mix}$.view($B$/2,$C$), $k$.view($C$,$B$)) \quad \textcolor{teal}{// semi-positive logits: [($B$/2)x$B$]}} \par
    \HiLi{\qquad $l_{neg}$ = torch.mm($q_{mix}$.view($B$/2,$C$), queue.view($C$,$K$)) \quad \textcolor{teal}{// negative logits: [($B$/2)x$K$]}} \par
    \medskip
    
    \HiLi{\qquad logits = torch.cat([$l_{semi\_pos}$, $l_{neg}$], dim=1) \quad \textcolor{teal}{// logits: [($B$/2)x($B$+$K$)]}} \par
    \HiLi{\qquad \textcolor{teal}{// $\lambda$ and (1-$\lambda$) for $i$-th and ($i$+$B$/2)-th dim, mixco labels: [($B$/2)x($B$+$K$)]}} \par
    \HiLi{\qquad labels = torch.cat([torch.diag($\lambda$, 0) + torch.diag((1-$\lambda$, $B$/2)), torch.zeros($K$)])} \par
    \HiLi{\qquad loss += torch.nn.KLDivLoss(logits / $\tau_{mix}$, labels)} \par
    \medskip
    
    \qquad loss.backward() \par
    \qquad $f_q$.params = torch.optim.SGD($f_q$.grads) \quad \textcolor{teal}{// SGD update for query network} \par 
    \qquad $f_k$.params = $m$ * $f_k$.params + (1-$m$) * $f_q$.params \quad \textcolor{teal}{// momentum update for key network} \par 
    \medskip
        
    \qquad queue = enqueue\_and\_dequeue(queue, $k$) \quad \textcolor{teal}{// maintain the queue size by $K$} \par
    \medskip

\end{algorithm}

%%%%%%%%%%%%%%%%%%%%%%%%%%%%%%%%%%%%%%%%%%%%%%%%%%%%%%%%%%%%%%%%%%%%%%%%%%%%%%%%%
\section{Memory and GPU time usage}
\label{appendix:computation}

\begin{table}[h!]
    \centering
    \begin{tabular}{lcccc}
    \toprule
    Architecture (epochs) & MixCo & Memory (GB/GPU) & Time (s/epoch) & Top-1 Accuracy (\%) \\
    \midrule
    ResNet-18 (100) & X & 3.5 & 132.6 & 35.79 \\
    ResNet-18 (100) & O & 4.5 & 153.2 & 42.65 \\
    ResNet-34 (200) & X & 4.3 & 246.3 & 42.15 \\
    ResNet-50 (200) & X & 8.1 & 306.8 & 45.82 \\
    \bottomrule
    \end{tabular}
    \vskip 0.1 in
    \caption{Required resources to train MoCo-v2 \citep{moco2} encoder. Every experiment is conducted on TinyImageNet with 2 RTX 2080Ti GPUs.}
    \label{tab:memory_computation}
\end{table}
In \autoref{tab:memory_computation}, we compare the required resources to pretrain MoCo-v2 encoder with MixCo pretraining. Using MixCo is more efficient in memory, computation, and time than vanilla MoCo-v2. MixCo requires more (mixed) inputs than MoCo, and the number of additional inputs is half the batch size according to our implementation. The increased number of inputs implies the increment of memory and the computation of logit vectors. As with the ResNet-18 encoder, this causes around $\times$1.29 memory and $\times$1.16 time cost. Compared with ResNet-34 encoder, ResNet-18 MixCo encoder produces similar test accuracy. However, ResNet-34 encoder requires more computation due to the larger number of parameters. Therefore, training time is $\times$3.22 larger than ResNet-18 MixCo (note that ResNet-34 is trained for 200 epochs to be comparable with ResNet-18 MixCo performance).

\section{Additional experiments}\label{appendix:additional}

\subsection{ImageNet experiment}
We compare MixCo to other methods in common ImageNet \citep{ILSVRC15} experiment protocol. MixCo achieved 68.4\% of linear classification accuracy, which is comparable to the state-of-the-art model. We set $\tau_{mix}=1$.
\vspace{-5pt}
\begin{table}[h!]
    \caption{ImageNet linear evaluation top-1 accuracy (1-crop, 224$\times$224), trained on features from the unsupervised pretraining. We compared several existing methods that share the similar settings: architecture, number of epochs, and batch size, since the performance highly depends on these factors.}
    \vspace{5pt}
    \label{tab:imagenet}
    \centering
    \begin{tabular}{llccc}
    \toprule
    Method & architecture & epochs & batch size & ImageNet acc. (\%) \\
    \midrule
    MoCo-v1 \citep{moco} & R50 & 200 & 256 & 60.6 \\
    PCL \citep{pcl} & R50 & 200 & 256 & 61.5 \\
    CPC-v2 \citep{cpc2} & R50 & 200 & 512 & 63.8 \\
    SimCLR \citep{simclr} & R50-MLP & 200 & 256 & 61.9 \\
    MoCo-v2 \citep{moco2} & R50-MLP & 200 & 256 & 67.5 \\
    PCL-v2 \citep{pcl} & R50-MLP & 200 & 256 & 67.6 \\
    SiMo \citep{eqco} & R50-MLP & 200 & 256 & \bf 68.5 \\
    MixCo (ours) & R50-MLP & 200 & 256 & 68.4 \\
    \bottomrule
    \end{tabular}
\end{table}

\subsection{Pretraining with longer epochs}
\vspace{-8pt}
\begin{table}[h!]
    \caption{Linear evaluation results (\%) while pretraining with different number of epochs.}
    \vspace{3pt}
    \label{tab:longer_epochs}
    \centering
    \begin{tabular}{cccccc}
    \toprule
    Architecture & Epochs & \multicolumn{2}{c}{MoCo-v2 \citep{moco2}} & \multicolumn{2}{c}{MixCo (ours)} \\
    & & Acc@1 & Acc@5 & Acc@1 & Acc@5 \\ 
    \midrule
    ResNet-18 & 100 & 35.79 & 61.16 & \bf 42.65 & \bf 67.77 \\
    ResNet-18 & 200 & 42.12 & 68.33 & \bf 45.15 & \bf 70.25 \\
    ResNet-18 & 400 & 45.73 & 72.17 & \bf 48.27 & \bf 72.72 \\
    \bottomrule
    \end{tabular}
\end{table}
\end{CJK}
\end{document}